\documentclass[10pt,twocolumn,letterpaper]{article}

\usepackage{wacv}
\usepackage{times}
\usepackage{epsfig}
\usepackage{graphicx}
\usepackage{color}
\usepackage{color, colortbl}
\usepackage{adjustbox}
\usepackage{array, enumitem, soul, algorithm, algpseudocode}
\usepackage{amsmath}
\usepackage{amssymb}
\usepackage{cite}
\usepackage{xcolor}
\usepackage{xspace}
\usepackage{booktabs}
\usepackage{pifont}
\usepackage{url}
\usepackage{comment}
\usepackage{multirow}
\usepackage[accsupp]{axessibility}  % Improves PDF readability for those with disabilities.

\definecolor{Gray}{gray}{0.95}
\newcommand{\cmark}{\text{\ding{51}}}%
\newcommand{\xmark}{\text{\ding{55}}}%

\makeatletter
\DeclareRobustCommand\onedot{\futurelet\@let@token\@onedot}
\def\@onedot{\ifx\@let@token.\else.\null\fi\xspace}

\def\ie{\emph{i.e}\onedot}

\def\et{\emph{et al}\onedot}

\makeatother

\def\wacvPaperID{} % Enter the WACV Paper ID here

\wacvapplicationstrack % Uncomment this line if you are submitting to the Applications Track.

\wacvfinalcopy % *** Uncomment this line for the final submission

\usepackage[pagebackref,breaklinks,colorlinks]{hyperref}
\usepackage[accsupp]{axessibility}  % Improves PDF readability for those with disabilities.

\begin{document}

%%%%%%%%% TITLE
\title{Perceptual Image Enhancement for Smartphone Real-Time Applications}

\author{Marcos V. Conde~$^{1}$, Florin Vasluianu~$^{1}$, Javier Vazquez-Corral~$^{2}$, Radu Timofte~$^{1}$\\
{\normalsize$^{1}$Computer Vision Lab, CAIDAS \& IFI, University of Würzburg, Germany}\\
{\normalsize$^{2}$Computer Vision Center and Computer Science Dept., Universitat Autònoma de Barcelona, Spain}\\
%$^{3}$Computer Vision Center, Campus UAB, Spain\\
{\tt\small \{marcos.conde-osorio,radu.timofte\}@uni-wuerzburg.de}\\
{\tt\normalsize {\url{https://github.com/mv-lab/AISP}}}
}

\maketitle
\thispagestyle{empty}

%%%%%%%%% ABSTRACT
\begin{abstract}
Recent advances in camera designs and imaging pipelines allow us to capture high-quality images using smartphones. However, due to the small size and lens limitations of the smartphone cameras, we commonly find artifacts or degradation in the processed images. The most common unpleasant effects are noise artifacts, diffraction artifacts, blur, and HDR overexposure. Deep learning methods for image restoration can successfully remove these artifacts. However, most approaches are not suitable for real-time applications on mobile devices due to their heavy computation and memory requirements. 

In this paper, we propose LPIENet, a lightweight network for perceptual image enhancement, with the focus on deploying it on smartphones. Our experiments show that, with much fewer parameters and operations, our model can deal with the mentioned artifacts and achieve competitive performance compared with state-of-the-art methods on standard benchmarks. Moreover, to prove the efficiency and reliability of our approach, we deployed the model directly on commercial smartphones and evaluated its performance. Our model can process 2K resolution images under 1 second in mid-level commercial smartphones.
\end{abstract}

%%%%%%%%% BODY TEXT

\section{Introduction}
\label{sec:intro}

\begin{figure}[!ht]
    \centering
    \includegraphics[width=\linewidth]{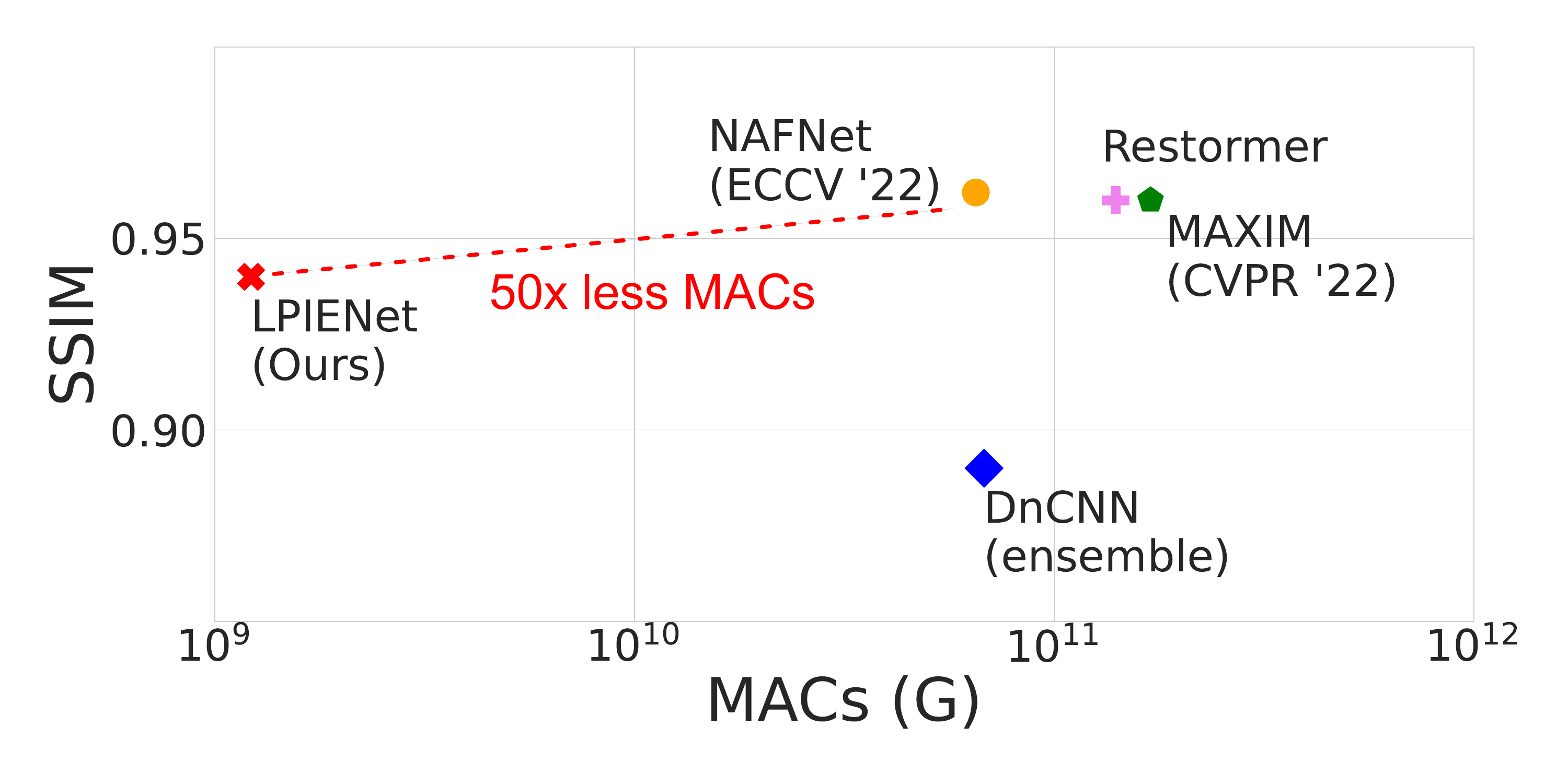}
    \caption{Comparison of computational cost and performance of \emph{state-of-the-art} methods for image denoising (SIDD)~\cite{abdelhamed2018, tu2022maxim, Zamir22, Wang_2022_CVPR, chen2022simple}. We can process 2K resolution images in 0.4s, and 4K images in 1.5 seconds on regular smartphone GPUs.}
    \label{fig:flops_params}
\end{figure}

In recent years the number of images that are captured has increased exponentially. The main reason for this surge comes from the ubiquitous presence of smartphones in our daily life. Phone manufacturers are continuously competing with the goal of delivering better images to their consumers in order to increase their sales. Therefore, a lot of research has been focused on improving the perceptual quality of these sRGB images.
%Therefore, a lot of research has been looking in how to obtain these better images.

Image restoration aims at improving the images captured by the cameras by removing different degradations introduced during image acquisition. These degradations can be introduced due to the physical limitations of cameras, for example the small aperture and limited dynamic range of smartphone cameras \cite{Zamir20}, or by inappropriate lighting conditions (\ie{} images captured in low-light). To solve these problems, image restoration is usually understood as an ill-posed problem, in which, given the degraded image the algorithm needs to output a clean image.

To be embedded in-camera by a manufacturer, an image restoration algorithm should comply with strong requirements in terms of quality, robustness, computational complexity, and execution time. In general, digital cameras have a set of resources in which to allocate all the operations in the ISP pipeline~\cite{conde2022modelbased}. Therefore, any new operation to be introduced in this pipeline should be of good enough quality to ``pay'' for the resources it will consume. Moreover, for an algorithm to be embedded in a camera, it is required to always improve over the input image, \ie{} to be robust for any possible circumstance and input signal. 

Image restoration is a traditional problem, and its study began as soon as we started to capture images, and many famous methods, such as Non-local Means for image denoising \cite{NLM}, are almost 20 years old. These traditional methods were usually defined by hand-crafted priors that narrowed the ill-posed nature of the problems by reducing the set of plausible solutions. However, since 2012 there has been a switch to deep learning based image restoration algorithms, as these methods have proven to be very powerful to generalize priors from a large number of images. 

Unfortunately, despite the great advances and performance, research on image restoration and enhancement using deep learning usually forgets the previous defined need for obtaining algorithms that have low computational complexity and execution time; and therefore many of them cannot be integrated into modern smartphones due to their complexity \ie{ FLOPs or memory requirements}. 

In this paper, we therefore aim at defining a new image enhancement algorithm that achieves competitive results in comparison with state-of-the-art methods on different related tasks, yet, at the same time, presents a low-complexity and a competitive execution time in current off-the-shelf smartphones as proven by the use of the AIScore~\cite{Ignatov2018AIBR}. A first example of this behaviour is shown in Figure~\ref{fig:flops_params}, where we compare our method to the current state-of-the-art in image denoising. As it can be seen, our methods is as close as 0.02 in SSIM to the state-of-the-art, while having at least $\times$50 less MACs. More details will appear later in Section~\ref{sec:eff}. %when discussing Table \ref{tab:eff-comp}.
\vspace{2mm}

\noindent In summary, \textbf{our contributions} are as follows:
\begin{itemize}
    \item We propose a lightweight U-Net based architecture characterized by the inverted residual attention (IRA) block. Similar to contemporary works in this field, yet more efficient and smaller.
    \item We optimize our model in terms of used parameters and computational cost (\ie{ FLOPs, MACs}), thus being able to achieve real-time performance on current smartphone GPUs at FullHD input image resolution. This improvement is illustrated in Figure~\ref{fig:flops_params}.
    \item We propose a new type of analysis, from a production point of view, observing the behaviour of our model when deployed on commercial smartphones. % hardware, targeting different market segments and various price levels.  
\end{itemize}

%\clearpage
%%%%%%%%%%%%%%%% RELATED WORK

\section{Related Work}
\label{sec:rel-work}

Image restoration is split in a large number of sub-problems, and in this paper we focus on four of the most popular in current research: image denoising, image deblurring, HDR image reconstruction from a single image, and Under-Display-Camera (UDC) image restoration.

\paragraph{Image denoising}
Image denoising has been a topic of research for more than 30 years. The most famous traditional image denoising methods are the non-local ones, such as Non-Local-Means \cite{NLM} and BM3D \cite{BM3D}. %, which are still competitive in current realistic datasets, such as SIDD \cite{abdelhamed2018}. 
More recently, multiple methods have studied different image representations to facilitate the denoising problem for this well-behaved algorithms \cite{Ghimpeteanu,VazquezCorral}.

As in other image restoration problems, research on image denoising has moved towards deep learning models. The first remarkable work on denoising with deep learning is probably Zhang \emph{et al.} \cite{Zhang-TIP} DnCNN, where they proposed to learn a CNN to estimate the noise distribution of the input image. Since then, plenty of other deep learning methods have appeared \cite{Zamir21,Zamir22,Wang2022,Anwar2019,Yue2019,Guo19,Liu2021,Chang2020,Zamir20}. This has also been possible thanks to the appearance of challenges such as \cite{abdelhamed2020ntire, abdelhamed2018} that facilitate a benchmark for compering methods together with training and testing images. For a deeper analysis we refer the reader to the survey in \cite{tian2020deep}.

\paragraph{Image deblurring}
Image deblurring is a traditional problem in image restoration. Its main objective is to remove the blur that appears in the input image, this can be caused by different factors (\ie{} camera shake, object motion, or lack of focus) and output a sharp image.
As it is the case for image denoising, initially the different algorithms were based on hand-made priors or constraints, mostly treating image restoration as an inverse filtering problem \cite{cho2009fast,fergus2006removing,xu2010two}, but these methods were surpassed with the appearance of deep learning models \cite{nah2017deep, tao2018scale,kupyn2019deblurgan,cho2021rethinking,tu2022maxim}. For a more in-depth analysis of deep learning methods applied to this problem, we point the reader to the survey in \cite{zhang2022deep}

\paragraph{HDR reconstruction}
Digital cameras can only capture around two orders of magnitude in luminance, and have therefore a very limited Dynamic Range. This is far beyond the luminance differences that appear in the real world, and the reason why, when capturing images, we may end up with highlights in very bright regions or information loss in very dark ones. 

The first works aiming at recovering the dynamic range of images were based on a set of multiple images from the same scene. The seminar work of Debevek and Malik \cite{DM} assumed that from a set of multiples images of the same scene, it is possible to recover a single Camera Response Function (CRF) and undo the camera process. This hypothesis was true for film cameras, but as recently proved by Gil Rodriguez \emph{et al.} \cite{GilRodriguez19} this is not true for current digital cameras, as color channels are not independent and the camera modifies the non-linearity for different exposure values.

Currently, thanks to deep learning and different benchmark challenges~\cite{perez2022ntire}, there has been a surge of methods aiming at recovering the full High Dynamic Range (HDR) from a single input image, a problem also named inverse tone mapping \cite{banterle2006inverse, banterle2009high}. This rising started with the work of Eilersten \emph{et al.} \cite{Eilertsen}, where they proposed a U-Net architecture for solving this problem. Other deep learning based methods for HDR single-image reconstruction are the ones in \cite{liu2020single,Santos2020}. Regarding deep learning methods for multiple images input, we should also mention the work in \cite{catley2022flexhdr}.

\begin{figure*}[!ht]
    \centering
    \includegraphics[width=\linewidth]{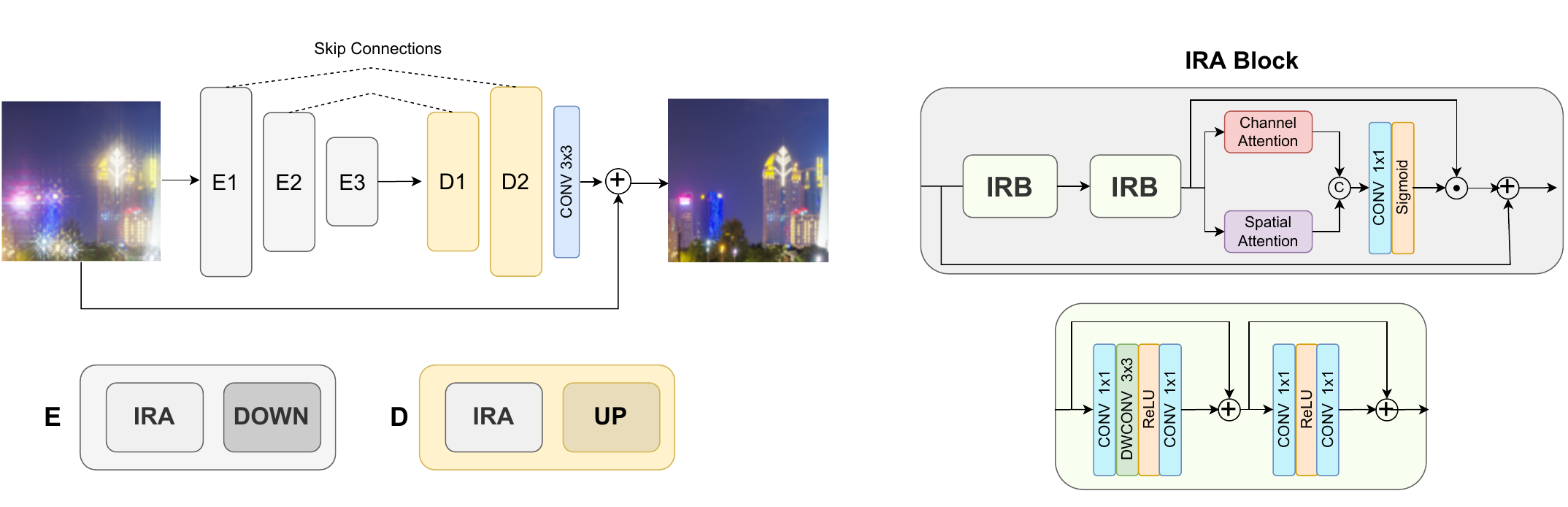}
    \caption{Architecture of the proposed \emph{LPIENet} network and the IRA Block. Our model is designed considering current \textsc{TFLite} supported operations and mobile devices limitations. The attention mechanisms~\cite{woo2018cbam, chen2022simple} have been optimized to require less memory and computation.}
    \label{fig:main}
\end{figure*}

\paragraph{UDC image restoration}
Recently, a new imaging system, the Under-Display Camera (UDC) has appeared. The UDC system consists of a camera module placed underneath and closely attached to the semi-transparent Organic Light-Emitting Diode (OLED) display~\cite{zhou2020udc}. This solution provides an advantage when it comes to the user experience analysis, with the full-screen design providing a higher level of comfort. The disadvantage of this solution is that the OLED display acts as an obstacle for the light interacting with the camera sensor, inducing additional reflections, refractions and other effects connected to the Image Signal Processing (ISP)~\cite{conde2022modelbased} model characterizing the camera. 

Zhou \et{}~\cite{udc-ir-cvpr21} and their 2020 ECCV challenge \cite{zhou2020udc} were the first works that directly addressed this novel restoration problem using deep learning. Baidu Research team~\cite{zhou2020udc} proposed the Residual dense based on Shade-Correction for T-OLED UDC Image Restoration. In \cite{udc-ir-cvpr21}, the authors devised an MCIS to capture paired images, and solve the UDC image restoration problem as a blind deconvolution problem. More recently, Feng \et{}~\cite{Feng_2021_CVPR_Discnet} proposed one of the world's first production UDC device for data collection, experiments, and evaluations. They also proposed a new model called DISCNet~\cite{Feng_2021_CVPR_Discnet}, and provided a benchmark for multiple blind and non-blind methods on UDC datasets.

\paragraph{General problem formulation}
It is important to note that the UDC problem can be seen as a generalization of the other three listed above. Following the formulation proposed in \cite{Feng_2021_CVPR_Discnet}, the UDC problem can be formulated as:
\begin{equation}
\label{eqn:formation}
    y = \gamma(C(x*k+n)),
\end{equation}
where $x$ represents the clean image with high dynamic range, $k$ is the point spread function (PSF) (i.e. the blurring kernel), $*$ represents the 2D convolution operator, and $n$ denotes the camera noise. Also, $C(\cdot)$ emulates the reduction of dynamic range, following $C(x)=\text{min}(x, x_{max})$, where $x_{max}$ is a range threshold, and $\gamma(\cdot)$ represents a Tone-Mapping Function.

From this formulation we can clearly see that when we suppose there is not noise ($n=0$ at all pixels), $x_{max}$ is large enough and we do not perform any Tone-Mapping, we end up with the traditional deblurring formulation $y=x*k$. Similarly, if we suppose that the kernel $k$ is a Dirac Delta, $x_{max}$ is equal to the maximum possible value of the input signal and we do not perform any Tone-Mapping we end up with the traditional image denoising formulation $y=x+n$. As we will be using in this paper the SIDD dataset~\cite{abdelhamed2018}, we will suppose that $n= \mathcal{N}(0, \beta)$, where $\beta^2(y) = \beta_1y + \beta_2$ and $\beta_1$ represents the shot noise and $\beta_2$ represents the independent additive Gaussian noise.
Finally, supposing that there is not noise ($n=0$ at all pixels), and  that the kernel $k$ is a Dirac Delta, we end up with the HDR image reconstruction problem.

\hspace{0.1cm}

However, as explained in the introduction, none of the aforementioned methods have analyzed these problems from the efficiency point of view. The proposed models can generate high quality results but cannot be integrated into modern smartphones due to their complexity \ie{ FLOPs}.

%\clearpage
%%%%%%%%%%%%%%%% PROPOSED METHOD

\section{Proposed Method}

We propose a new model called \emph{LPIENet}, following a U-Net~\cite{ronneberger2015unet} like architecture, standard in image restoration~\cite{Wang2022, zhang2022deep, Feng_2021_CVPR_Discnet, udc-ir-cvpr21, Zamir22}. The main building blocks are inverted residual attention blocks, we refer to them as IRA blocks. These are selected due to their efficiency~\cite{howard2017mobilenets}. This is a blind image restoration method, and therefore, we do not rely on the PSF or other information about the camera sensor. This architecture is illustrated in Figure~\ref{fig:main}. The initial model consists on 5 blocks (3 encoders and 2 decoders) with $[16, 32, 64, 32, 16]$ channels respectively, and 0.13M parameters. 

To prove the scalability of our approach we design a deeper network by increasing the number of channels to $[32, 64, 128, 64, 32]$. We refer to this modification as \emph{LPIENet-L}.
Our base model represents a solution with $5\times$ less parameters than other methods~\cite{Wang2022, Feng_2021_CVPR_Discnet, udc-ir-cvpr21, chen2022simple, Zamir22}. Note that \emph{LPIENet} was tailored specifically to be efficient when deployed on various smartphone devices. As this model is designed to perform denoising, deblurring and UDC image restoration in real-time conditions on mobile devices, we optimize it considering SSIM performance and a lower number of FLOPs. As we will prove in Section~\ref{sec:eff}, this model achieves competitive performance while processing Full-HD and 2K resolution images in real-time on a wide range of commercial smartphones.
Moreover, as we pointed out in Section~\ref{sec:rel-work}, our methods combine ideas from \textbf{deblurring}~\cite{nah2021ntire} and \textbf{HDR}~\cite{liu2020singlehdr, catley2022flexhdr} networks, and attention methods~\cite{woo2018cbam, Zamir22}. We believe these tasks are extremely correlated with the UDC restoration problem.

\subsection{Method Description}

As we show in Figure~\ref{fig:main}, \emph{LPIENet} consists of: Three encoder blocks ($E_1$, $E_2$, $E_3$) and two decoder blocks ($D_1$, $D_2$). Each encoder block consists of the following sequence, what we previously named an IRA block: two inverted linear residual blocks~\cite{howard2017mobilenets}, an attention block~\cite{woo2018cbam} to enrich the feature representation, and finally, a downsampling layer (\ie{} max-pooling). 
The decoder blocks $D_1$, $D_2$ follow the same structure. We replace the downsampling operations by a bilinear upsampling layer, and concatenate the skip-connections~\cite{ronneberger2015unet}. An important detail to improve efficiency is the fact that we upsample the features after activating them, not before.
Since the inverted residual blocks~\cite{howard2017mobilenets} have a limited receptive field and representation, we further activate them using a combination of spatial and color attention~\cite{woo2018cbam}. 

Contemporary work by Chen \et{} has shown this is an efficient image restoration solution~\cite{chen2022simple}. Other important considerations are: (i) we do not use Batch Normalization (BN) layers, which consume the same amount of GPU memory as convolutional layers, and they also increase computational complexity~\cite{zhang2018residualdense}, (ii) \texttt{LayerNorm} and \texttt{GELU} have shown a consistent performance improvement in image restoration~\cite{chen2022simple}, however, due to the deployment limitations we do not use such techniques. (iii) We report MACs or FLOPs considering the following standard relationship between both terms: $\text{MACs} \approx 0.5 \times \text{FLOPs}$

\section{Experimental Setup}

\subsection{Datasets}

\textbf{SIDD Medium}~\cite{abdelhamed2018, abdelhamed2020ntire} is a real image denoising dataset providing 320 image pairs, two image pairs from each one of the 160 scenes. The dataset introduces great variability in terms of sensors, with five different smartphone camera sensors used for data acquisition, and images captured under different exposure and lighting conditions. A noise model estimation method is then used to produce the counterparts of the noise-affected images.
The authors provide $\approx$ 80\% of the data for training and validation, with the remaining 20\% not being publicly available. Methods are compared and tested using their on-line SIDD benchmark~\footnote{\url{https://www.eecs.yorku.ca/~kamel/sidd/benchmark.php}} and test set of 1280 images.
We further processed the high-resolution training images (\ie{ $3000\times2000$}) to a set of non-overlapping image crops, extracting 12 crops of resolution $1000\times1000$ per each training image.

\vspace{2mm}

\textbf{UDC SYNTH}~\cite{Feng_2021_CVPR_Discnet, feng2022mipi}. The original RGB images from SYNTH~\cite{Feng_2021_CVPR_Discnet} have resolution $800\times800$, we extracted non-overlapping patches of size $400\times400$. We extracted patches (instead of downsampling the image) to preserve the high-frequency details. Moreover, images were transformed from the original domain to the tone-mapped domain using the following function $f(x) = x / (x+0.25)$. Therefore the pixel intensities are in the range $[0,1)$.

\vspace{2mm}

\textbf{GoPro}~\cite{nah2017deep} by Nah \et{} is widely used in motion deblurring, for training and evaluation.
It consists of 3214 pairs of blurry and sharp HD resolution images ($1280\times720$). This is a synthetic dataset since the blurred images are produced by averaging several high-speed clean images. Following the standard practice~\cite{nah2017deep}, we use 2103 pairs for training and 1111 pairs for testing. During training we use random paired crops of size $540\times960$.

\subsection{Implementation details}

To avoid artifacts, the output of our models is clipped into range $[0,1-p]$ , where $p$ is $10^{-5}$. Our models were implemented in Tensorflow 2 and trained using a TPU v3.

We used Adam optimizer~\cite{kingma2014adam} with default hyper-parameters, and an initial learning rate of 2e-3. The learning rate is reduced by 50\% during plateaus up to a minimum learning rate of $1e^{-6}$. We use basic augmentations: horizontal and vertical flip, and rotations. We set 4 as mini-batch size and trained to convergence for a few days (\ie{} 500 epochs). We found very profitable to train on large resolution, we start training with random crops of size $400\times400$ en eventually increase the image size up to HD resolution.

\vspace{-4mm}
\paragraph{Loss function}

The model is trained using a weighted sum of a $\mathcal{L}_1$ loss, a SSIM loss and a Gradient loss~\cite{ma2020structure}. The final loss function is:
\begin{equation}
\label{eqn:7}
    \mathcal{L} = \alpha \mathcal{L}_{\text{SSIM}}+\mathcal{L}_1+ \beta \mathcal{L}_{\text{Grad}}
\end{equation}

where $\alpha$ and $\beta$ are set empirically to scale the loss.

\subsection{Experimental Results}

\begin{table}[t]%[!ht]
    \centering
    \caption{Quantitative Results for Under-display Camera (UDC) Image Restoration using SYNTH~\cite{Feng_2021_CVPR_Discnet}. We show fidelity and perceptual metrics. (*) indicates methods proposed at the MIPI UDC Challenge~\cite{feng2022mipi}. We consult some numbers from~\cite{Feng_2021_CVPR_Discnet, feng2022mipi}. We highlight in bold the best blind and non-blind methods.
    \vspace{0.2cm}}
    \label{tab:results}
    
    \resizebox{\linewidth}{!}{
    \begin{tabular}{l c c c c c}
         %Method & PSNR$\uparrow$ &\multicolumn{2}{c||}{Perc. metrics} &\# Params & Extra\\
         %name &  (dB) & SSIM$\uparrow$ & LPIPS$\downarrow$ & (milions) & data \\
         %\hline\noalign{\smallskip}
         \toprule
         
         Method & PSNR~(dB)~$\uparrow$ & SSIM~$\uparrow$ & LPIPS~$\downarrow$  &\#~Param.~(M) & PSF \\
         %name &  (dB) & SSIM$\uparrow$ & LPIPS$\downarrow$ & (milions) & data \\
         \midrule
         
         Wiener Filter (WF)~\cite{orieux2010bayesian-wiener} & 27.30 & 0.83 & 0.330 & - & \cmark \\
         SRMDNF~\cite{Feng_2021_CVPR_Discnet} & 34.80 & 0.96 & 0.036 & 1.5 & \cmark \\
         SFTMD~\cite{Feng_2021_CVPR_Discnet} & 42.35 & 0.98 & 0.012 & 3.9 & \cmark \\
         \textbf{DISCNet (PSF)}~\cite{Feng_2021_CVPR_Discnet} & \textbf{42.77} & \textbf{0.98} & 0.012 & 3.8 & \cmark \\
         %\hline
         RDUNet~\cite{zhang2018residualdense, ronneberger2015unet}& 34.37 & 0.95 & 0.040 & 8.1 & \xmark\\
         DE-UNet~\cite{udc-ir-cvpr21} & 38.11 & 0.97 & 0.021 & 9.0 & \xmark \\
         DISCNet (w/o PSF)~\cite{Feng_2021_CVPR_Discnet} & 38.55 & 0.97 & 0.030 & 2.0 & \xmark \\
         %\hline
         RushRushRush~* & 39.52 & 0.98 & 0.021 & n/a & \xmark\\
         eye3~* & 36.69 & 0.97 & 0.032 & n/a & \xmark\\
         FMS Lab~* & 35.77 & 0.97 & 0.045 & n/a & \xmark\\
         EDLC2004~* & 35.50 & 0.96 & 0.045 & n/a & \xmark\\
         SAU\_LCFC~* & 32.75 & 0.96 & 0.056 & n/a & \xmark\\
         \midrule
         \rowcolor{Gray} \textbf{\emph LPIENet-L (ours)} & \textbf{40.12} & \textbf{0.98} & 0.020 & 0.6 & \xmark\\
         \rowcolor{Gray} \textbf{\emph LPIENet (ours)} & 34.10 & 0.95 & 0.031 & 0.1 & \xmark\\
         \bottomrule
    \end{tabular}}
\end{table}

%\subsubsection{Quantitative Results}

\paragraph{UDC image restoration} We evaluate our models on the SYNTH dataset~\cite{Feng_2021_CVPR_Discnet}, also used in the ``UDC MIPI 2022 Challenge"~\cite{feng2022mipi}, and compare them against current state-of-the-art methods (excluding contemporary methods and methods that have not been officially published).
Note that these results are not fully reproducible since multiple approaches do not provide open-sourced code. Moreover, the SYNTH dataset~\cite{Feng_2021_CVPR_Discnet} test set (and PSF) is public, which makes difficult a fair comparison with methods that might tend to overfit it.
As we show in Table~\ref{tab:results}, we achieve competitive results for this problem, while using models with fewer parameters than the other methods. As an example, we can look at the comparison against DiscNet~\cite{Feng_2021_CVPR_Discnet} base blind version, as we used the same initial setup to build up our model. We can see how our method outperforms this one by almost 2 dBs. We extend this analysis and prove the benefits of our method in Section~\ref{sec:eff}.

\vspace{-3mm}
\paragraph{Real image denoising} In Table \ref{sidd-benchmark} we present our results for the SIDD~\cite{abdelhamed2018}. As we can see our method is quite competitive, getting as close as 0.02 in SSIM to the state-of-the-art. We should recap here the analysis shown in Figure~\ref{fig:flops_params}, were we can see that our methods requires 100x less MACs than Restormer~\cite{Zamir22} and MAXIM~\cite{tu2022maxim} (we will see the difference in MACs in the next section). This allows our method to be deployed in current smartphones (as we will show next in Section~\ref{sec:eff}), while Restormer is far from that capability. Please remind that this was the goal of our paper: to obtain a lightweight model able to compete with state-of-the-art, but with reduced runtime and complexity, so it is possible to embed the model in current smartphones.
We provide our architecture ablation study in Table~\ref{tab:ablation}. The attention mechanisms~\cite{chen2022simple} help to improve the performance of the efficient inverted residual blocks.

\begin{table}[t]
   \centering
   \caption{Quantitative sRGB Denoising results on the SIDD~\cite{abdelhamed2018}. Our model outperforms classical famous methods and can denoise images in real-time in smartphones. Numbers consulted at~\cite{abdelhamed2018, chen2022simple}.
   \vspace{0.01cm}
   }
   \label{sidd-benchmark}
   \resizebox{0.84\linewidth}{!}{
   \begin{tabular}{l c c}
   \toprule
   Method & PSNR~(dB)~$\uparrow$ & SSIM~$\uparrow$ \\
   \midrule
   Noisy & 23.70 & 0.480 \\
   BM3D \cite{BM3D} & 25.65 & 0.685  \\
   %NLM \cite{NLM} & 26.75 & 0.699  \\
   KSVD \cite{ksvd} & 26.88 & 0.842  \\
   FoE \cite{foe} & 25.58 & 0.792  \\
   MLP \cite{mlp} & 24.71 & 0.641  \\
   WNNM \cite{wnnm} & 25.78  & 0.809  \\
   %
   %GLIDE \cite{glide} & 24.71 & 0.774  \\
   TNRD \cite{tnrd} & 24.73  & 0.643 \\
   EPLL \cite{epll} & 27.11  & 0.870 \\
   DnCNN \cite{Zhang-TIP} & 30.71 & 0.695  \\
   CBDNet \cite{guo2019toward}&	33.28 	&0.868 	\\
   CycleISP \cite{Zamir_2020_CVPR} & 39.52 & 0.957 \\
   DAGL \cite{mou2021dynamic} & 38.94 & 0.953 \\
   Mirnet \cite{Zamir20} & 39.73 & 0.959 \\
   Restormer \cite{Zamir22} & 40.02 & 0.960 \\
   \midrule
   \rowcolor{Gray} \textbf{\emph LPIENet (ours)} & 37.47 & 0.940  \\
   \rowcolor{Gray} \textbf{\emph LPIENet (ensemble)} & 37.73 & 0.943  \\
   \bottomrule
   \end{tabular}}
\end{table}

\begin{table}[t]
    \centering
    \caption{Ablation study for Real Image Denoising. SSIM reported on SIDD~\cite{abdelhamed2018}. GMACs calculated using a $256\times256$ RGB input.
    \vspace{0.2cm}
    }
    \resizebox{\linewidth}{!}{
    \begin{tabular}{l c c}
         \toprule
         Method & SSIM~$\uparrow$ & GMACs~$\downarrow$ \\
         \midrule
         Baseline UNet~\cite{Feng_2021_CVPR_Discnet, ronneberger2015unet} & 0.855 & 5.8 \\ 
         Res Block $\xrightarrow{}$ Dual Inv Res Block & 0.840 & 1.0 \\ 
         \rowcolor{Gray} Dual Inv Block $\xrightarrow{}$ IRA Block (Ours) & 0.940 & 1.3 \\ 
         LPIENet + kernel size = 5 & 0.941 & 1.4 \\
         \bottomrule
    \end{tabular}
    }
    \label{tab:ablation}
\end{table}

\begin{table*}[t]
    \centering
    \caption{Description of the selected commercial smartphone devices.}
    \vspace{0.1cm}
    \label{tab:phones}
    \resizebox{\linewidth}{!}{
    \begin{tabular}{l l l c l c c}
        \toprule
        Phone Model & Launch & Chipset & CPU & GPU & RAM (GB) & AI Score~$\uparrow$ \\
        \midrule
        (\# 1) Samsung A50 & 03/2019 & Exynos 9610 & 8 cores & Mali-G71 GP3 & 4 & 45.4  \\
        (\# 2) OnePlus Nord 2 5G & 07/2021 & MediaTek Dimensity 1200 & 8 cores & Mali-G77 MC9 & 8 & 194.3 \\
        (\# 3) OnePlus 8 Pro & 04/2020 & Qualcomm Snapdragon 865 5G & 8 cores & Adreno 650 & 12 & 137.0 \\
        (\# 4) Realme 8 Pro & 03/2021 & Qualcomm Snapdragon 720G & 8 cores & Adreno 618 & 8 & 60.6 \\
        \bottomrule
    \end{tabular}
    }
\end{table*}

\subsubsection{Qualitative Results}
\begin{figure}[!t]
\begin{center}
\scriptsize
\begin{tabular}[t]{c@{ }c@{ }c@{ }c} \hspace{-2mm}
\includegraphics[width=.11\textwidth]{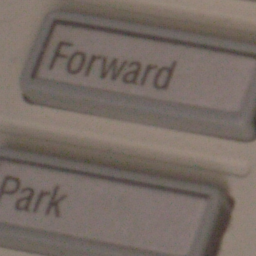}&    \hspace{-1.2mm}
\includegraphics[width=.11\textwidth]{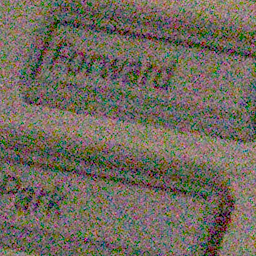}& \hspace{-1.2mm}
\includegraphics[width=.11\textwidth]{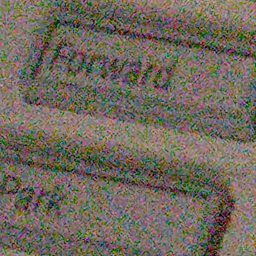}&   \hspace{-1.2mm}
\includegraphics[width=.11\textwidth]{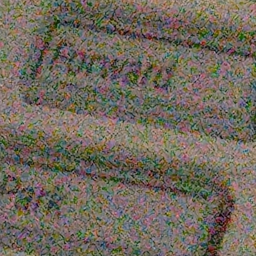}\\
 & 18.25 dB & 19.70 dB& 20.76 dB \\
Reference & Noisy & FFDNet~\cite{zhang2018ffdnet}& DnCNN~\cite{zhang2017beyond} \\ \hspace{-2mm}
\includegraphics[width=.11\textwidth]{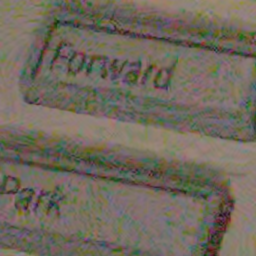}&  \hspace{-1.2mm}
\includegraphics[width=.11\textwidth]{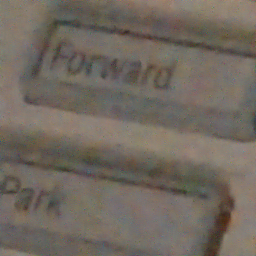}&  \hspace{-1.2mm}
\includegraphics[width=.11\textwidth]{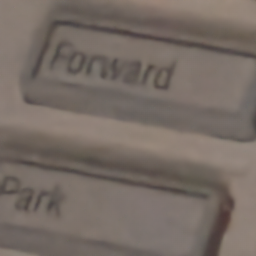}&  \hspace{-1.2mm}
\includegraphics[width=.11\textwidth]{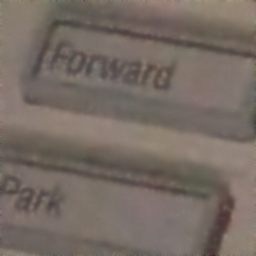}\\
25.75 dB& 28.84 dB & 35.57 dB& \textbf{33.42dB}\\
BM3D~\cite{BM3D}  & CBDNet~\cite{Guo19}  & RIDNet~\cite{Anwar2019}& \textbf{LPIENet (Ours)} \\

\includegraphics[width=0.11\textwidth]{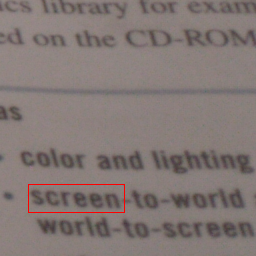} &
\includegraphics[width=0.11\textwidth]{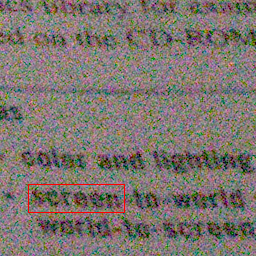} &
\includegraphics[width=0.11\textwidth]{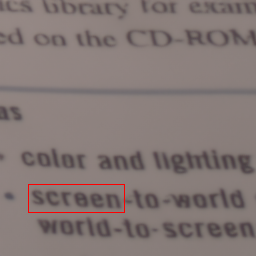} &
\includegraphics[width=0.11\textwidth]{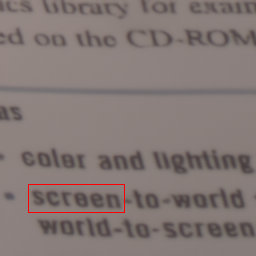} \\
\includegraphics[width=0.11\textwidth]{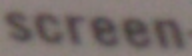} &
\includegraphics[width=0.11\textwidth]{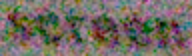} &
\includegraphics[width=0.11\textwidth]{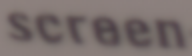} &
\includegraphics[width=0.11\textwidth]{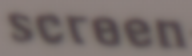} \\
 & 19.01 dB & 35.21 dB& ~35.01 dB \\ 
Reference & Noisy & ~HINet\cite{Chen_2021_CVPR} & ~Restormer\cite{Zamir22} \\ 
\includegraphics[width=0.11\textwidth]{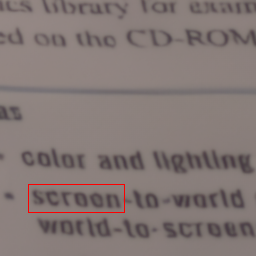} &
\includegraphics[width=0.11\textwidth]{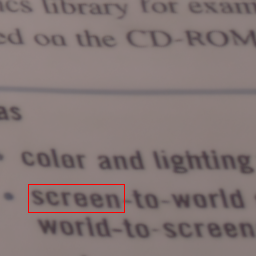} &
\includegraphics[width=0.11\textwidth]{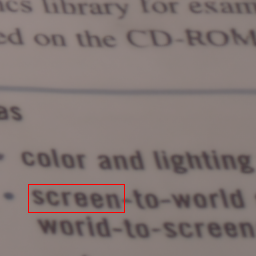} &
\includegraphics[width=0.11\textwidth]{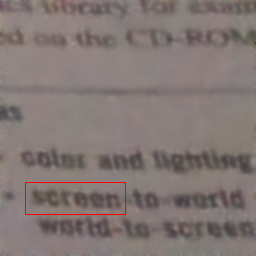} \\
\includegraphics[width=0.11\textwidth]{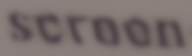} &
\includegraphics[width=0.11\textwidth]{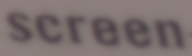} &
\includegraphics[width=0.11\textwidth]{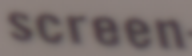} &
\includegraphics[width=0.11\textwidth]{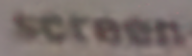} \\
 ~34.96 dB & ~35.97 dB & ~35.77 dB  & 33.57dB\\ 
~MPRNet\cite{Zamir21} & NAFNet-baseline\cite{chen2022simple} & NAFNet \cite{chen2022simple} & \textbf{LPIENet (ours)} \\
\end{tabular}
\end{center}
%\vspace*{-6mm}
\caption{Real image denoising results of different methods on a challenging sRGB image from the SIDD dataset~\cite{abdelhamed2018}. Our method can recover details and produce pleasant results while being $10\times$ smaller than other methods.}
\label{fig:srgb_sidd}
%\vspace*{-2mm}
\end{figure}

\paragraph{Real image denoising} Example results for the real image denoising case are shown in Figure~\ref{fig:srgb_sidd}, where we can clearly see that our results are competing again the state-of-the-art at a fraction of computational cost. We recommend the reader to focus on the words ``Forward" and ``Park'' in the top image, and in the word ``screen" in the bottom image. Once again please remember that our model has \emph{at least 50x less MACs} than any of the methods outrunning it. In Section~\ref{sec:eff}, we deploy our method and CycleISP \cite{Zamir_2020_CVPR} in commercial smartphones, and compare runtimes.

\vspace{-3mm}
\paragraph{HDR image reconstruction} Example results for the case of HDR image reconstruction are shown in Figure~\ref{fig:dicnet-comp}, we can see how our method is able to better reconstruct the textures, colors and geometric properties of the hallucinated objects, in comparison versus the other ones, that are state-of-the-art. More in detail, we want the reader to focus on the leaves of the plant in the top image, and the light specularities in the bottom one. Please also note that these images are from the SYNTH dataset~\cite{Feng_2021_CVPR_Discnet}, because, as we explained in the related work, UDC image restoration generalises the problem of HDR reconstruction. 

\vspace{-3mm}
\paragraph{Image deblurring} We process images from the GoPro dataset~\cite{nah2017deep} test split. We improve the base blurry inputs by 2dBs using our method (\ie{ average PSNR of 27.8 dB in the test set}). We also outperform baseline methods Xu \et{}~\cite{Xu_unnaturall0}, Hyun \et{}~\cite{6751504}, Whyte \et{}~\cite{5540175} and Gong \et{}~\cite{Gong2017FromMB}. 
%
%We also outperform baseline methods such as \cite{Xu_unnaturall0, 6751504, 5540175, Gong2017FromMB}. 
We provide visual results in Figure~\ref{fig:gopro} and in the supplementary material.

We also acknowledge other state-of-the-art methods such as NAFNet\cite{chen2022simple}, UFormer\cite{Wang_2022_CVPR}, Restormer\cite{Zamir22}, MAXIM\cite{tu2022maxim} and MPRNet\cite{Zamir21}, tested on the same deblurring and denoising setup, however, we do not compare with them as many are contemporary and more complex models that achieve the most competitive performance at the cost of being extremely inefficient. We analyze this in Table \ref{tab:eff-comp}.

%%%%%%% EFFICIENCY

\section{Efficiency Analysis}
\label{sec:eff}

We evaluate the performance of different methods using four smartphone devices.
In Table~\ref{tab:phones}, we provide all the details regarding the devices used in our benchmark. %We choose three different smartphones that target the \emph{entry-level} segment of the  market (1,2,4), and we also consider a mobile phone designed for the \emph{flagship} segment (3). %To observe the evolution of the entry-level smartphones in terms of AI deployment capabilities, we compare the Samsung A50 to the OnePlus Nord 2 5G. 
As a reference metric to measure the performance of each mobile device in the scenario of the deep learning models deployment, we provide the AI Score~\cite{Ignatov2018AIBR}. 
This AI score is computed over a set of experiments (\ie{ category recognition, semantic segmentation,or super-resolution}), where inference precision and hardware/software behaviour are observed and quantified into a score metric depending on the importance of each of the factors.
All the tests are described by the authors in \cite{Ignatov2018AIBR}, and are available for the Android smartphones through the application \emph{AI Benchmark} \cite{Ignatov2018AIBR}, which also provides support for deep learning models deployment on different hardware options (CPU, GPU, NPU), and with various precision types (\ie{} FP16, FP32, INT8). We refer to the work~\cite{Ignatov2018AIBR} for more details about this app. % the architecture and software details of this app.

\vspace{-3mm}
\paragraph{Deployment Limitations} Before introducing our benchmark setup and compared methods, we must note the limitations of this efficiency analysis: (i) models need to be converted to \textsc{TFLite} format, this conversion is not trivial and might lead to loss of performance, especially if the model was originally implemented using PyTorch (\ie{ \textsc{PT} $\xrightarrow{}$ \textsc{ONNX} $\xrightarrow{}$ \textsc{TF} $\xrightarrow{}$ \textsc{TFLite}}) . (ii) Multiple state-of-the-art operations such as \texttt{GELU} activation or \texttt{LayerNorm}~\cite{chen2022simple} are not currently supported. For these reasons, multiple methods cannot be currently deployed using the standard setup~\cite{Ignatov2018AIBR} or require a Tensorflow implementation. We adapt our models to follow this standard~\cite{Ignatov2018AIBR} and re-implement (if possible) other methods.

\begin{figure}[!ht]
    \centering
    \includegraphics[width=0.925\linewidth]{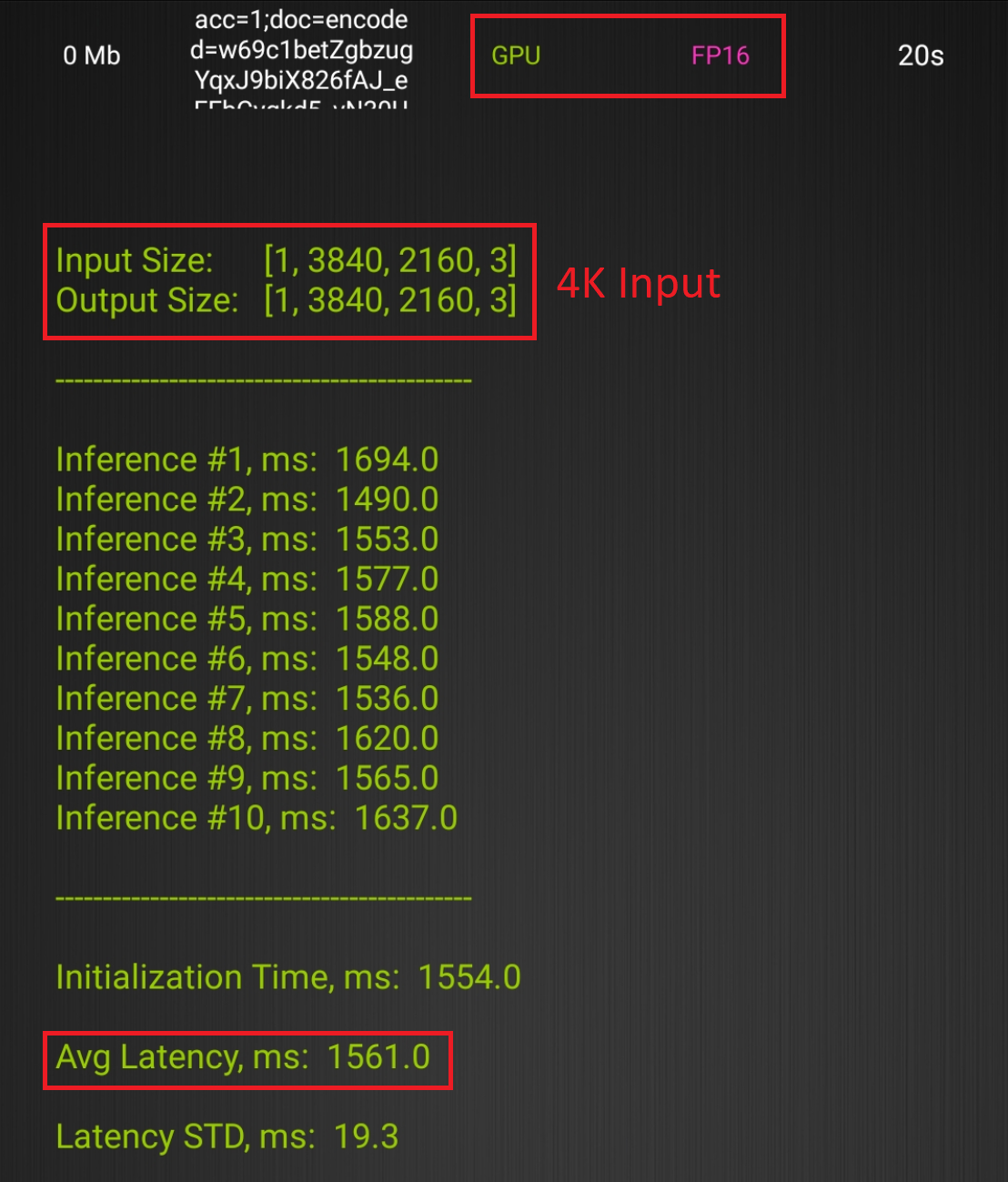}
    \vspace{1mm}
    \caption{\textbf{\emph AI Benchmark}~\cite{Ignatov2018AIBR} model performance evaluation. Our method can process 4K inputs in 1.5 seconds using GPU FP16.}
    \label{fig:aibench}
\end{figure}

\vspace{-3mm}
\paragraph{Benchmark setup} Models are tested on CPU and GPU (TFLite Delegate) FP16 after default Tensorflow-lite conversion and optimization, using as input a tensor image of different resolutions. The FLOPs are calculated for each resolution. We show in Figure~\ref{fig:aibench} the evaluation process using \emph{AI Benchmark app}~\cite{Ignatov2018AIBR}.

\begin{table}[t]
    \centering
    \caption{Comparison between \emph{LPIENet} and other state-of-the-art methods, in terms of perceptual performance (SSIM) and necessary mult-adds (MACs). ``Task" indicates the task the compared solutions were proposed for: denoising (1), deblurring (2), HDR imaging (3), and UDC camera restoratiron (4). ``Res" specifies the input resolution used to report the number of GMACs. The last column specifies their advantage in terms of perceptual performance (SSIM) over our model.}
    \vspace{0.2cm}
    \resizebox{\linewidth}{!}{
    \begin{tabular}{l c c c c}
         \toprule
         Method  & Task &  $\downarrow$MACs~(G) & Res. & SSIM Diff. \\
         \midrule
         MPRNet\cite{Zamir21} & 1/2 & 588 & 256 & 0.15 \\
         SRMDNF\cite{zhang2018learning} & 4 & 475.5 & 800 & 0.02 \\
         DISCNet~\cite{Feng_2021_CVPR_Discnet} & 4 & 221 & 800 & 0.01 \\
         HINet\cite{Chen_2021_CVPR} & 1/2 & 170.7 & 256 & 0.15\\
         MAXIM\cite{tu2022maxim} & 1/2 & 169.5 & 256 & 0.17 \\
         Restormer\cite{Zamir22} & 1/2 & 140 & 256 & 0.17\\
         UFormer\cite{Wang_2022_CVPR} & 1/2 & 89.5 & 256 & 0.17\\
         DE-UNet~\cite{udc-ir-cvpr21} & 4 & 84.5 & 800 & -0.01\\
         HDR-NTIRE~\cite{perez2022ntire} & 3 & $\leq$66 & FHD & -  \\
         NAFNet\cite{chen2022simple} & 1/2 &  65 & 256 &  0.24  \\
         \midrule
         \rowcolor{Gray} \textbf{\emph LPIENet (ours)} & 1/2/3/4 & \textbf{12} & 800 & - \\
         \rowcolor{Gray} \textbf{\emph LPIENet (ours)} & 1/2/3/4 & \textbf{1.3} & 256 & - \\
         \bottomrule
    \end{tabular}
    }
    \label{tab:eff-comp}
\end{table}

\vspace{-3mm}
\paragraph{Methods comparison} We use DiscNet ``baseline (a)" defined at~\cite{Feng_2021_CVPR_Discnet} to evaluate the model, and obtain a fair lower-bound of the complete method's performance. 
SRMDNF~\cite{udc-ir-cvpr21} and SFTMD~\cite{udc-ir-cvpr21} cannot be directly converted either, therefore, we use a canonical UNet model with the same number of FLOPs to approximate their performance. As we show in Table~\ref{tab:bench}, our proposed model \emph{LPIENet} can process Full-HD images in real-time in commercial mobile devices GPUs, being $\times5$ faster than DiscNet~\cite{Feng_2021_CVPR_Discnet}. 

\begin{figure*}[!ht]
    \centering
    \setlength{\tabcolsep}{2.0pt}
    \begin{tabular}{c}
    \includegraphics[width=\linewidth]{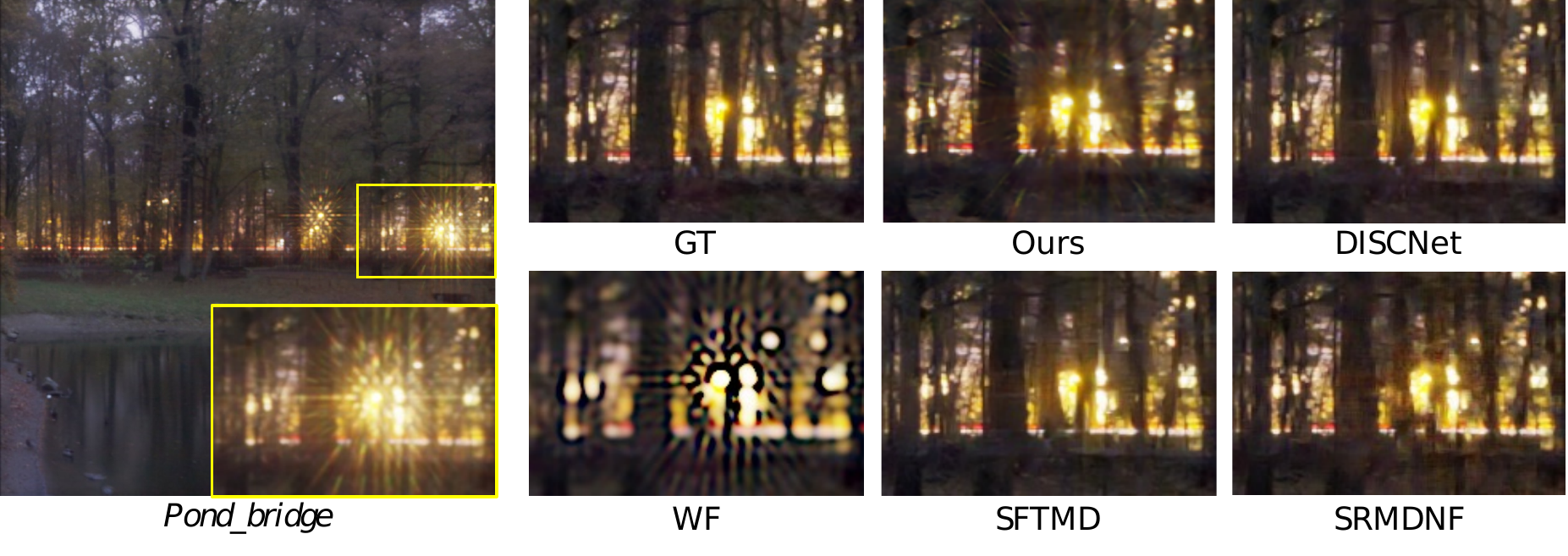}\tabularnewline
    \end{tabular}
    \caption{Visual comparison on \textbf{\emph SYNTH}~\cite{Feng_2021_CVPR_Discnet} synthetic validation images. Our method \emph{LPIENet-L} recovers fine details, and produces high perceptual quality images without unpleasant artifacts. Image selection from DiscNet~\cite{Feng_2021_CVPR_Discnet}.}
    \label{fig:dicnet-comp}
\end{figure*}

Our model can also process low-resolution images on CPU in real-time (under 1 second per image), and its performance only decays $0.04$ with respect to DiscNet~\cite{Feng_2021_CVPR_Discnet} in terms of SSIM~\cite{SSIMPaper}. 
In Figure~\ref{fig:dicnet-comp} we provide qualitative samples in challenging scenarios, our model is able to improve notably the perceptual quality of the UDC degraded images in real-time.
Note that in the perception-distortion tradeoff~\cite{blau2018perception}, we consider more important perceptual metrics~\cite{gu2022ntire, conde2022conformer}, as we aim for pleasant results for humans.

\vspace{-3mm}
\paragraph{Limitations} Besides the mentioned implementation limitations that are limiting performance (\ie{} \texttt{LayerNorm} is not supported, and was proven to be essential in image restoration~\cite{chen2022simple}), we find the following limitations in our method: (i) the use of pixel-wise convolutions an inverted residual blocks limits the receptive field, this is a clear drawback when solving HDR and deblurring problems (\ie{} the model cannot hallucinate and generate realistic content as more complex methods), (ii) the self-imposed limitation of operations is an evident performance limitation.

%%%%%%%% CONCLUSION

\section{Conclusions}
In this paper, we have presented a new lightweight model for perceptual image enhancement, called LPIENet. Our experiments and tests prove its ability to compete against the state-of-the-art in different image restoration and enhancement tasks, namely image denoising, image deblurring, HDR reconstruction and UDC image restoration. Our new model consists of only 0.1M parameters and is able to process high resolution images in current off-the-shelf phones in less than a second.

The main novelty of our new model is on the combination of a U-Net like architecture with inverted residual attention (IRA) blocks that allows to drastically reduce the number of operations required.
We are able to challenge well-established models with a fraction of their reported number of parameters or number of performed operations. Finally, we introduce a novel benchmark for image enhancement based on efficiency and deployment capabilities on real smartphones. We refer the reader to our project page for supplementary material.

\vspace{4mm}
\small{
\noindent\textbf{Acknowledgments} This work was partly supported by the The Alexander von Humboldt Foundation (AvH).\\
JVC was supported by Grant PID2021-128178OB-I00 funded by MCIN/AEI/10.13039/501100011033 and by ERDF "A way of making Europe", and also by the "Ayudas para la recualificación del sistema universitario español" financed by the European Union-NextGenerationEU.
}

\begin{table*}[!h]
    \centering
    \caption{\textbf{Efficiency benchmark for UDC Image Restoration on different commercial smartphones.} We show the performance of our method in terms of runtime at different image resolutions, device architectures and running scenarios (CPU, GPU). Runtimes are the average of at least 5 iterations. Our method is the one of the few methods described in this work that can process high-resolution images, even 2K, without tiling or patching the input image, while we achieving high perceptual quality results. Note that the methods marked with $\xmark$ failed the test due to memory requirements or runtime constraints. FHD is considered $1920\times1080$.}
    \vspace{0.15cm}
    \label{tab:bench}
    \resizebox{\linewidth}{!}{
    \begin{tabular}{l|c|c|c||c|c||c|c||c|c||c|c}
        \hline\noalign{\smallskip}
        & & &  &\multicolumn{6}{c}{Runtime (s)} \\
        Method & FLOPs~$\downarrow$ & Resolution &  SSIM~$\uparrow$ &\multicolumn{2}{c||}{Phone \#1} & \multicolumn{2}{c||}{Phone \#2} & \multicolumn{2}{c||}{Phone \#3} & \multicolumn{2}{c}{Phone \#4}\\
         & (G) & (px) &  & CPU & GPU & CPU & GPU & CPU & GPU & CPU & GPU \\
         \hline
         SFTMD~\cite{udc-ir-cvpr21} & 2460 & 800$\times$800 & 0.986 & \xmark & \xmark & \xmark & \xmark & \xmark & \xmark & \xmark & \xmark \\
         \hline
         \rowcolor{Gray} RDUNet~\cite{yang2020rdunet} & 2461 & FHD & \multirow{3}{*}{0.970} & $\xmark$ & 21.2  & 437.8 & 2.7 & $\xmark$ & 5.1 & 764.5 &$\xmark$ \\
         RDUNet~\cite{yang2020rdunet} & 759 & 800$\times$800 & & 261.1 & 6.2  & 65.4 & 0.86 & 137.1 & 1.50 & 113.9 & 4.49 \\
         RDUNet~\cite{yang2020rdunet} & 78 & 256$\times$256  & & 1.4 & 0.6  & 0.39 & 0.09 & 0.58 & 0.18 & 0.64 & 0.52\\
         \hline
         \rowcolor{Gray} CycleISP~\cite{Zamir_2020_CVPR} & 12410 & FHD & \multirow{3}{*}{0.957} & $\xmark$ & 98.7  & $\xmark$ & 20.51 & $\xmark$ & $\xmark$ & \xmark & \xmark \\
         CycleISP~\cite{Zamir_2020_CVPR} & 3830 & 800$\times$800 &  & $\xmark$ & 29.4  & 844.9 & 5.9 & $\xmark$ & 10.17 & 1264  & $\xmark$ \\
         CycleISP~\cite{Zamir_2020_CVPR} & 392 & 256$\times$256 &  & 14.6 & 3.0  & 5.8 & 0.64 & 8.2 & 1.0 & 7.54 & 3.2\\
         \hline
         \rowcolor{Gray} DiscNet~\cite{Feng_2021_CVPR_Discnet} & 1434 & FHD & \multirow{3}{*}{0.974} & 1055 & 13.7  & 261.5 & 2.0 & 61.1 & 5.3 & 455.2 & \xmark \\
         DiscNet ~\cite{Feng_2021_CVPR_Discnet} & 442 & 800$\times$800 &  & 52.4 & 4.0 & 13.5 & 0.63 & 15.6 & 1.4 & 23.3 & 3.98  \\
         DiscNet  ~\cite{Feng_2021_CVPR_Discnet} & 256 & 256$\times$256 & & 1.1 & 0.41 & 0.3 & 0.07 & 0.6 & 0.14 & 0.58 & 0.38\\
         \hline
         \rowcolor{Gray} LPIENet & 310.4 & 4K & & $\xmark$ & 6.9 & 368.1 & 1.5 & 73.0 & 1.6  & $\xmark$  & $\xmark$ \\
         \rowcolor{Gray} LPIENet & 82.8 & 2K &  & 24.8 & 1.8 & 6.9 & 0.43 & 13.1 & 0.4 & 12.40 & 1.2\\
         \rowcolor{Gray} LPIENet & 77.6 & FHD & 0.940 & 22.0 & 1.7 & 6.3 & 0.41 & 12.0 & 0.4 &11.57 & 1.15 \\
         LPIENet & 23.95 & 800$\times$800 & & 3.8 & 0.52 & 1.5 & 0.13 & 2.2 & 0.13 &2.4 & 0.372 \\
         LPIENet & 2.45 & 256$\times$256 &  & 0.38 & 0.065 & 0.14 & 0.031 & 0.23 & 0.015 & 0.26 & 0.044 \\
         %\hline
         \hline
    \end{tabular}}
\end{table*}

\begin{figure*}[!ht]
    \centering
    \setlength{\tabcolsep}{1.0pt}
    \begin{tabular}{ccc}
    \includegraphics[width=0.315\linewidth]{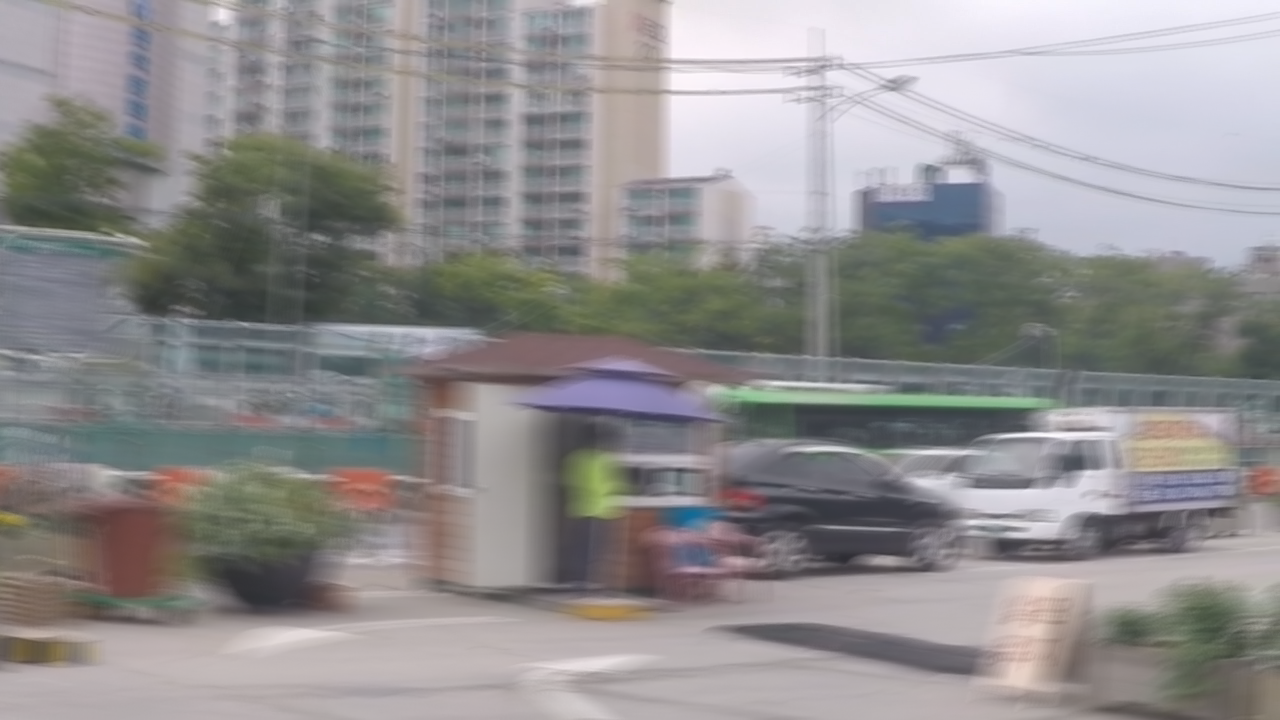} &
    \includegraphics[width=0.315\linewidth]{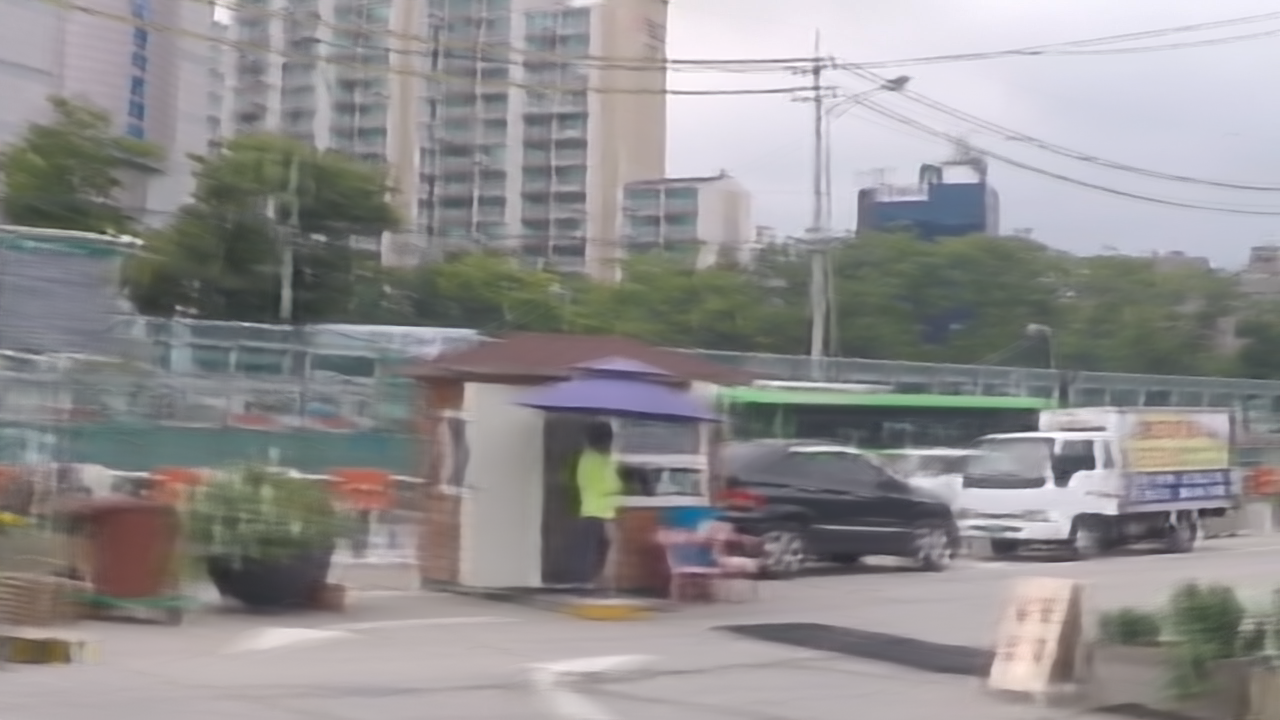} &
    \includegraphics[width=0.315\linewidth]{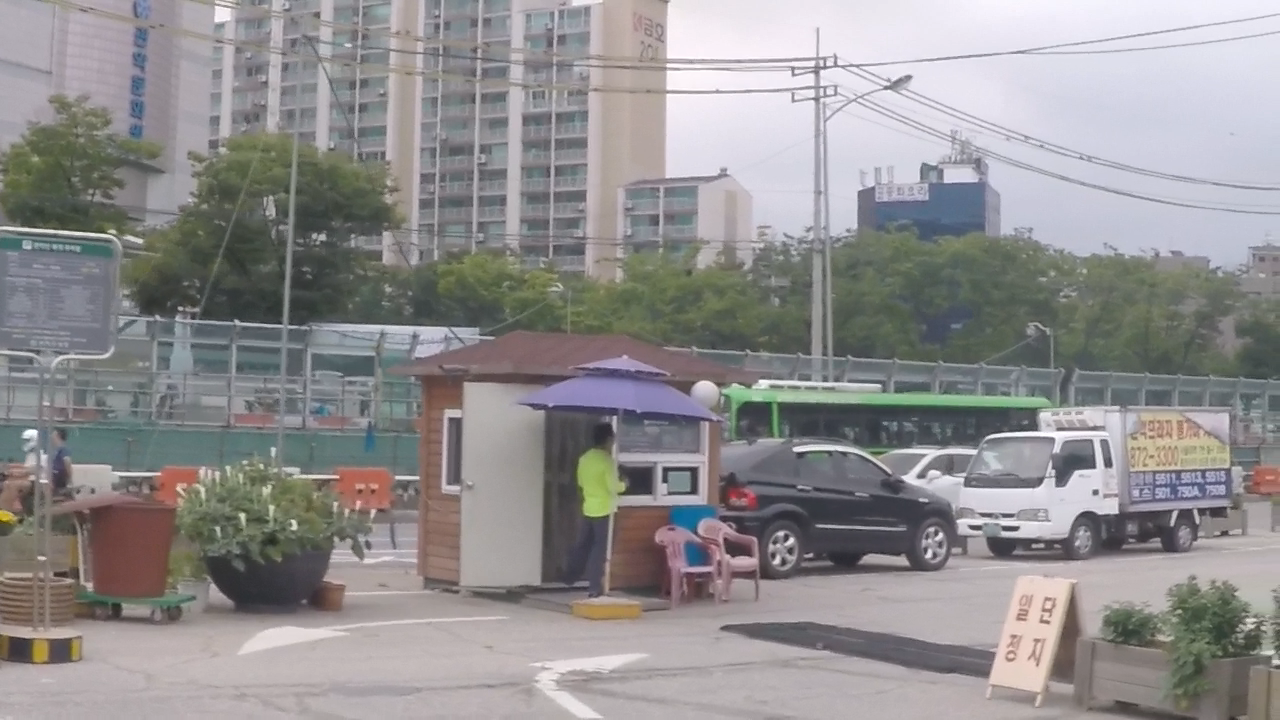}
    \tabularnewline
    Input & \emph{LPIENet (ours)} & Reference
    \end{tabular}
    \vspace{2mm}
    \caption{Qualitative results on the \textbf{\emph GoPro dataset}~\cite{nah2017deep} test split. Our model can also reduce notably motion blur.}
    \label{fig:gopro}
\end{figure*}

{\small
\bibliographystyle{ieee_fullname}
\bibliography{egbib}
}

\end{document}